\pdfoutput=1
\documentclass[11pt,a4paper]{article}
\usepackage{tikz}
\usepackage{lipsum,adjustbox}
\usepackage{amsmath}
\usepackage[hyphens]{url}
\usepackage[hyperref]{emnlp2018}
\usepackage{times}
\usepackage{latexsym}
\usepackage{array}
\usepackage{scrextend}
\usepackage{url}

\usepackage[utf8]{inputenc}

\DeclareFixedFont{\ttb}{T1}{txtt}{bx}{n}{10} 
\DeclareFixedFont{\ttm}{T1}{txtt}{m}{n}{10}  

\usepackage{color}
\definecolor{deepblue}{rgb}{0,0,0.5}
\definecolor{deepred}{rgb}{0.6,0,0}
\definecolor{deepgreen}{rgb}{0,0.5,0}

\usepackage{listings}

\newcommand\HASH{\text{H}}

\newcommand\bashstyle{\lstset{
language=bash,
basicstyle=\ttm,
showstringspaces=false,
commentstyle=\ttb\color{gray},
otherkeywords={python,pip,pip2,pip3},
keywordstyle=\ttb\color{deepblue},
emph={-m, -o, -i},
emphstyle=\ttb\color{deepred}, 
stringstyle=\color{deepgreen},
frame=tb,                         
showstringspaces=false            %
}}

\lstnewenvironment{bash}[1][]
{
\bashstyle
\lstset{#1}
}
{}

\newcommand\bashinline[1]{{\bashstyle\lstinline!#1!}}

\newcommand\pythonstyle{\lstset{
language=Python,
basicstyle=\tiny\ttm,
otherkeywords={self},             
keywordstyle=\ttb\color{deepblue},
commentstyle     = \ttb\color{gray},
emph={Magnitude,FeaturizerMagnitude,__init__},          
emphstyle=\ttb\color{deepred},    
stringstyle=\color{deepgreen},
frame=tb,                         
showstringspaces=false            %
}}

\lstnewenvironment{python}[1][]
{
\pythonstyle
\lstset{#1}
}
{}

\newcommand\pythoninline[1]{{\pythonstyle\lstinline!#1!}}

\aclfinalcopy 

\title{Magnitude: A Fast, Efficient Universal Vector Embedding Utility Package }

\author{Ajay Patel \\
  Plasticity Inc.\\
  San Francisco, CA \\
  {\tt ajay@plasticity.ai}\\\And
  Alexander Sands \\
  Plasticity Inc.\\
  San Francisco, CA \\
  {\tt alex@plasticity.ai} \\\AND
  \textbf{Chris Callison-Burch} \\
  Computer and Information \\ Science Department\\
  University of Pennsylvania \\
  {\tt ccb@upenn.edu} \\\And
  \textbf{Marianna Apidianaki} \\
  LIMSI, CNRS\\
  Universit\'e Paris-Saclay \\
  91403 Orsay, France\\
  {\tt marapi@seas.upenn.edu} \\
 }

\date{}

\begin{document}
\maketitle
\begin{abstract}
Vector space embedding models like word2vec, GloVe, fastText, and ELMo are extremely popular representations in natural language processing (NLP) applications.  We present Magnitude, a fast, lightweight tool for utilizing and processing embeddings.  Magnitude is an open source Python package with a compact vector storage file format that allows for efficient manipulation of huge numbers of embeddings.  Magnitude performs common operations up to 60 to 6,000 times faster than Gensim. Magnitude introduces several novel features for improved robustness like out-of-vocabulary lookups.
\end{abstract}

\section{Introduction}
Magnitude is an open source Python package developed by Ajay Patel and Alexander Sands \cite{ajay_patel_2018_1255637}.  It provides a full set of features and a new vector storage file format that make it possible to use vector embeddings in a fast, efficient, and simple manner. It is intended to be a simpler and faster alternative to current utilities for word vectors like Gensim~\cite{rehurek_lrec}.

Magnitude's file format (``.magnitude") is an efficient universal vector embedding format.  The Magnitude library implements on-demand lazy loading for faster file loading, caching for better performance of repeated queries, and fast processing of bulk key queries.  Table~\ref{speed-ratios} gives speed benchmark comparisons between Magnitude and Gensim for various operations on the Google News pre-trained word2vec model~\cite{DBLP:journals/corr/abs-1301-3781}.  Loading the binary files containing the word vectors takes Gensim 70 seconds, versus 0.72 seconds to load the corresponding Magnitude file, a 97x speed-up.  Gensim uses 5GB of RAM versus 18KB for Magnitude. 

Magnitude implements functions for looking up vector representations for misspelled or out-of-vocabulary words, quantization of vector models, exact and approximate similarity search, concatenating multiple vector models together, and manipulating models that are larger than a computer's main memory. 
Magnitude's ease of use and simple interface combined with its speed, efficiency, and novel features make it an excellent tool for cases ranging from applications used in production environments to academic research to students in natural language processing courses.

\begin{table}
\centering
\scalebox{0.9}{
\begin{tabular}{|l|r|r|}
\hline
Metric & Cold & Warm \\
\hline
Initial load time & 97x & --\\
Single key query & 1x & 110x\\
Multiple key query (n=25)& 68x & 3x\\
k-NN search query (k=10) & 1x & 5,935x\\
\hline
\end{tabular}
}
\caption{Speed comparison of Magnitude versus Gensim for common operations.  The `cold' column represents the first time the operation is called. The `warm' column indicates a subsequent call with the same keys.}\label{speed-ratios}
\end{table}

\section{Motivation}

Magnitude offers solutions to a number of problems with current utilities.

\paragraph{Speed:} Existing utilities are prohibitively slow for iterative development.  Many projects use Gensim to load the Google News word2vec model directly from a ``.bin" or ``.txt" file multiple times.  It can take between a minute to a minute and a half to load the file.

\paragraph{Memory:} A production web server will run multiple processes for serving requests. Running Gensim, in the same configuration, will consume $>$4GB of RAM usage per process.

\paragraph{Code duplication:}
Many developers duplicate effort by writing commonly used routines that are not provided in current utilities. Namely, routines for concatenating embeddings, bulk key lookup, out-of-vocabulary search,  and building indexes for approximate k-nearest neighbors.

The Magnitude library uses several well-engineered libraries to achieve its performance improvements.  It uses SQLite\footnote{\url{https://www.sqlite.org/}} as its underlying data store, and takes advantage of database indexes for fast key lookups and memory mapping.  It uses NumPy\footnote{\url{http://www.numpy.org/}} to achieve significant performance speedups over native Python code using computations that follow the Single Instruction, Multiple Data (SIMD) paradigm.  It uses spatial indexes to perform fast exact similarity search and Annoy\footnote{\url{https://github.com/spotify/annoy}} to perform approximate k-nearest neighbors in the vector space. To perform feature hashing, it uses xxHash\footnote{\url{https://xxhash.org/}}, an extremely fast non-cryptographic hash algorithm, working at speeds close to RAM limits.  Magnitude's file format uses LZ4 compression\footnote{\url{http://www.lz4.org/}} for compact  storage.

\section{Design Principles}

Several design principles guided the development of the Magnitude library:
\begin{itemize}
\item The API should be intuitive and beginner friendly.  It should have sensible defaults instead of requiring configuration choices by the user. The option to configure every setting should still be provided to power users.
\item The out of the box configuration should be fast and memory efficient for iterative development.  It should be suitable for deployment in a production environment. Using the same configuration in development and production reduces bugs and makes 
deployment easier.
\item The library should use lazy loading whenever possible to remain fast, responsive, and memory efficient during development.
\item The library should aggressively index, cache, and use memory maps to be fast, responsive, and memory efficient for production.
\item The library should be able to process data that is too large to fit into a computer's main memory.
\item The library should be thread-safe and employ memory mapping to reduce duplicated memory resources when multiprocessing.
\item The interface should act as a generic key-vector store and remain agnostic to underlying models (like word2vec, GloVe, fastText, and ELMo) and remain useable for other domains that use vector embeddings like computer vision~\cite{babenko2016efficient}.
\end{itemize}

Gensim offers several speed ups of its operations, but these are largely only accessible through advanced configuration. For example, by re-exporting a ``.bin", ``.txt", or ``.vec" file into its own native format that can be memory-mapped. Magnitude makes this easier by providing a default configuration and file format that requires no extra configuration to make development and production workloads run efficiently out of the box.

\section{Getting Started with Magnitude}
The system consists of a Python 2.7 and Python 3.x compatible package (accessible through the PyPI index\footnote{\url{https://pypi.org/project/pymagnitude/}} or GitHub\footnote{\url{https://github.com/plasticityai/magnitude}}) with utilities for using the ``.magnitude" format and converting to it from other popular embedding formats.
\subsection{Installation}
Installation for Python 2.7 can be performed using the {\small\tt{pip}} 
command:
\begin{bash}
pip install pymagnitude
\end{bash}
Installation for Python 3.x can be performed using the {\small\tt{pip3}} 
command:
\begin{bash}
pip3 install pymagnitude
\end{bash}

\subsection{Basic Usage}
Here is how to construct the Magnitude object, query for vectors, and compare them:

\begin{python}
from pymagnitude import *
vectors = Magnitude("w2v.magnitude")
k = vectors.query("king")
q = vectors.query("queen")
vectors.similarity(k,q) # 0.6510958
\end{python}

Magnitude queries return almost instantly and are memory efficient.  It uses lazy loading directly from disk, instead of having to load the entire model into memory. Additionally, Magnitude supports nearest neighbors operations, finding all words that are closer to a key than another key, and analogy solving (optionally with \newcite{levy2014linguistic}'s 3CosMul function):

\begin{python}
vectors.most_similar(k, topn=5) 
#[(`king', 1.0), (`kings', 0.71),
# (`queen', 0.65), (`monarch', 0.64),
# (`crown_prince', 0.62)]

vectors.most_similar(q, topn=5) 
#[(`queen', 1.0), (`queens', 0.74),
#(`princess', 0.71), (`king', 0.65),
# ('monarch', 0.64)]

vectors.closer_than("queen", "king") 
#[`queens', `princess']

vectors.most_similar(
	positive = ["woman", "king"], 
	negative = ["man"]
) # queen
vectors.most_similar_cosmul(
	positive = ["woman", "king"], 
	negative = ["man"]
) # queen
\end{python}

\noindent In addition to querying single words, Magnitude also makes it easy to query for multiple words in a single sentence and multiple sentences:
\begin{python}
vectors.query("play") 
# Returns: a vector for the word
vectors.query(["play", "music"])
# Returns: an array with two vectors
vectors.query([
 ["play", "music"],
 ["turn", "on", "the", "lights"],
]) # Returns: 2D array with vectors
\end{python}


\subsection{Advanced Features}

\paragraph{OOVs:} 
Magnitude implements a novel method for handling out-of-vocabulary (OOV) words.  OOVs frequently occur in real world data since pre-trained models are often missing slang, colloquialisms, new product names, or misspellings. For example, while {\it uber} exists in Google News word2vec, {\it uberx} and {\it uberxl} do not. These products were not available when Google News corpus was built. Strategies for representing these words include generating random unit-length vectors for each unknown word or mapping all unknown words to a token like ``UNK'' and representing them with the same vector. These solutions are not ideal as the embeddings will not capture semantic information about the actual word. Using Magnitude, these 
OOV words can be simply queried and will be positioned in the vector space close to other 
OOV words based on their string similarity:
\begin{python}
"uber" in vectors # True
"uberx" in vectors # False
"uberxl" in vectors # False
vectors.query("uberx") 
  # Returns: [ 0.0507, -0.0708, ...]
vectors.query("uberxl")
  # Returns: [ 0.0473, -0.08237, ...]
vectors.similarity("uberx", "uberxl") 
  # Returns: 0.955
\end{python}

\noindent A consequence of generating 
OOV vectors is that misspellings and typos are also sensibly handled:
\begin{python}
"missispi" in vectors # False
"discrimnatory" in vectors # False
"hiiiiiiiiii" in vectors # False
vectors.similarity(
  "missispi",
  "mississippi"
) # Returns: 0.359
vectors.similarity(
  "discrimnatory",
  "discriminatory"
) # Returns: 0.830
vectors.similarity(
  "hiiiiiiiiii",
  "hi"
) # Returns: 0.706
\end{python}

\noindent The OOV handling is detailed in Section~\ref{sec:oov}.

\paragraph{Concatenation of Multiple Models:}
Magnitude makes it easy to concatenate multiple types of vector embeddings  to create combined models.

\begin{python}
w2v = Magnitude("w2v.300d.magnitude")
gv = Magnitude("glove.50d.magnitude")
vectors = Magnitude(w2v, gv) # concat
vectors.query("cat") 
# Returns: 350d NumPy array 
# 'cat' from w2v and 'cat' from gv
vectors.query(("cat", "cats")) 
# Returns: 350d NumPy array 
# 'cat' from w2v and 'cats' from gv
\end{python}

\paragraph{Adding Features for Part-of-Speech Tags and Syntax Dependencies to Vectors:} 

Magnitude can directly turn a set of keys (like a POS tag set) into vectors. Given an approximate upper bound on the number of keys and 
a namespace, it uses the hashing trick~\cite{DBLP:journals/corr/abs-0902-2206} to create an appropriate length dimension for the keys.
\begin{python}
pos_vecs = FeaturizerMagnitude(
  100, namespace = "POS")
pos_vecs.dim # 4
# number of dims automatically
# determined by Magnitude from 100
pos_vecs.query("NN")
dep_vecs = FeaturizerMagnitude(
  100, namespace = "Dep")
dep_vecs.dim # 4
dep_vecs.query("nsubj")
\end{python}

\noindent This can be used with Magnitude's concatenation feature to combine the vectors for words with the vectors for POS tags or dependency tags. Homonyms show why this may be useful:
\begin{python}
vectors = Magnitude(vecs, pos_vecs, 
                    dep_vecs)
vectors.query([
    ("Buffalo", "JJ", "amod"), 
    ("buffalo", "NNS", "nsubj"), 
    ("Buffalo", "JJ", "amod"), 
    ("buffalo", "NNS", "nsubj"), 
    ("buffalo",  "VBP", "rcmod"),
    ("buffalo",  "VB", "ROOT"),
    ("Buffalo",  "JJ", "amod"),
    ("buffalo",  "NNS", "dobj")
  ]) # array of 8 x (300 + 4 + 4)
\end{python}

\paragraph{Approximate k-NN}
We support approximate similarity search with the {\small\tt{most\_similar\_approx}} function. This finds the approximate nearest neighbors more quickly than the exact nearest neighbors search performed by the {\small\tt{most\_similar}} function. The method accepts an {\small\tt{effort}} argument which accepts the range $[0.0, 1.0]$. A lower {\small\tt{effort}} will reduce accuracy, but increase speed. A higher {\small\tt{effort}} does the reverse. This trade-off works by searching more- or less-indexed trees. Our approximate k-NN is powered by Annoy, an open source library released by Spotify.
%
Table~\ref{approximate-speeds} compares the speed of various configurations for similarity search.

\begin{table}
\centering
\scalebox{0.9}{
\begin{tabular}{|l|r|}
\hline
Metric & Speed \\
\hline
Exact k-NN  & 0.9155s \\
Approx. k-NN (k=10, effort = 1.0)  & 0.1873s \\
Approx. k-NN (k=10, effort = 0.1)  & 0.0199s\\
\hline
\end{tabular}
}
\caption{ Approximate nearest neighbors significantly speeds up similarity searches compared to exact search. Reducing the amount of allowed effort further speeds the approximate k-NN search. }\label{approximate-speeds}
\end{table}

\newcommand\CNGRAM{\text{CGRAM}}
\newcommand\PRVG{\text{PRVG}}
\newcommand\seed{\text{seed}}
\newcommand\BASIC{\text{B}}
\newcommand\ADVANCED{\text{A}}
\newcommand\boov{\text{oov}}
\newcommand\BOOV{\text{OOV}}
\newcommand\BOW{\text{BOW}}
\newcommand\EOW{\text{EOW}}
\newcommand\MATCH{\text{MATCH}}
\newcommand\aoov{\text{oov}}
\newcommand\AOOV{\text{OOV}}

\section{Details of OOV Handling}
\label{sec:oov}

Facebook's fastText~\cite{DBLP:journals/corr/BojanowskiGJM16} provides similar OOV functionality to Magnitude's.  Magnitude allows for OOV lookups for any embedding model, including older models like word2vec and GloVe~\cite{DBLP:journals/corr/abs-1301-3781,pennington2014glove}, which did not provide OOV support. Magnitude's OOV method can be used with existing embeddings because it does not require any changes to be made at training time like fastText's method does. For ELMo vectors, Magnitude will use ELMo's OOV method.

\paragraph{Constructing vectors from character n-grams:} 

We generate a vector for an OOV word $w$ based on the character n-gram sequences in the word.  First, we pad the word with a character at the beginning of the word and at the end of the word.
~Next, we generate the set of all character-ngrams in $w$ (denoted with the fuction  $\CNGRAM_{w}$) between length 3 and 6, following \newcite{DBLP:journals/corr/BojanowskiGJM16}, although these parameters are tunable arguments in the Magnitude converter.
We use the set of character n-grams $C$ to construct a vector $\BOOV_d(w)$ with $d$ dimensions to represent the word $w$.  Each unique character n-gram $c$ from the word contributes to the vector through a {\bf p}seudo{\bf r}andom {\bf v}ector {\bf g}enerator function \PRVG.  Finally, the vector is normalized. 
\begin{align*}
C & = \CNGRAM_{w}(3,6) \\
\boov_d(w) & = \sum_{c~\in~C}\PRVG_{\HASH(c)}(-1.0, 1.0, d)\\
\BOOV_d(w) & = \frac{\boov_d(w)}{\lvert\boov_d(w)\rvert}
\end{align*}

\noindent \PRVG's random number generator is seeded by the value ``$\seed$", which generates uniformly random vectors of dimension size $d$, with values in the range  of -1 to 1. 
The hashing function $\HASH$ produces a 32 bit hash of its input using xxHash. $\HASH:  \{0,1\}^* \rightarrow \{0,1\}^{32}$.
Since the $\PRVG$'s seed is only conditioned upon the word $w$, the output is deterministic across different machines. 

This character n-gram-based method will generate highly similar vectors for a pair of OOVs with similar spellings, like {\it uberx} and {\it uberxl}.  However, they will not be embedded close to similar in-vocabulary words like {\it uber}.  


\paragraph{Interpolation with in-vocabulary words}

To handle matching OOVs to in-vocabulary words, we first define a function $\MATCH_{k}(a,b,w)$.  $\MATCH_{k}(a,b,w)$ returns the normalized mean of the vectors of 
the top $k$ most string-similar in-vocabulary words using the full-text SQLite index. In practice, we use the top 3 most 
string-similar words. These are then used to interpolate the values for the vector representing the OOV word.  30\% of the weight for each value comes from the pseudorandom vector generator based on the OOV's n-grams, and the remaining 70\% comes from the values of the 3 most string similar in-vocabulary words: 
\begin{align*}
\aoov_d(w) & = [0.3 * \BOOV_d(w) \\ 
&~~+ 0.7 * \MATCH_{3}(3,6,w)]
\end{align*}

\paragraph{Morphology-aware matching}

For English, we have implemented a nuanced string similarity metric that is prefix- and suffix-aware.  While {\it uberification} has a high string similarity to {\it verification} and has a lower string similarity to {\it uber}, good OOV vectors should weight stems more heavily than suffixes.
Details of our morphology-aware matching are omitted for space.

\paragraph{Other matching nuances}
We employ other techniques when computing the string similarity metric, such as shrinking repeated character sequences of three or more to two ({\it hiiiiiiii} $\rightarrow$ {\it hii}), ranking strings of a similar length higher, and ranking strings that share the same first or last character higher for shorter words.


\section{File Format}
To provide efficiency at runtime, Magnitude uses a custom ``.magnitude" file format instead of ``.bin", ``.txt", ``.vec", or ``.hdf5'' that word2vec, GloVe, fastText, and ELMo use~\cite{DBLP:journals/corr/abs-1301-3781,pennington2014glove,DBLP:journals/corr/JoulinGBM16,DBLP:journals/corr/abs-1802-05365}. 
%
The ``.magnitude" file is a SQLite database file.  There are 3 variants of the file format: Light, Medium, Heavy. Heavy models have the largest file size but support all of the Magnitude library's features. Medium models support all features except approximate similarity search. Light models do not support approximate similarity searches or interpolated 
OOV lookups, but they still support basic 
OOV lookups. See Figure~\ref{file-format} for more information about the structure and layout of the ``.magnitude" format.

\tikzset{every picture/.style={line width=0.75pt}} 
\begin {figure}[!hbtp]
\centering
\scalebox{.5}{
\begin{adjustbox}{width=\textwidth}
\begin{tikzpicture}[x=0.75pt,y=0.75pt,yscale=-.7,xscale=1]

\draw    (82, 122) rectangle (379.5, 162)   ;
\draw    (239.5,194.42) -- (238.75,165) ;
\draw [shift={(238.75,165)}, rotate = 448.54] [color={rgb, 255:red, 0; green, 0; blue, 0 }  ]   (0,0) .. controls (3.31,-0.3) and (6.95,-1.4) .. (10.93,-3.29)(0,0) .. controls (3.31,0.3) and (6.95,1.4) .. (10.93,3.29)   ;

\draw    (81, 227) rectangle (396.5, 267)   ;
\draw    (240.5,302.42) -- (239.75,273) ;
\draw [shift={(239.75,273)}, rotate = 448.54] [color={rgb, 255:red, 0; green, 0; blue, 0 }  ]   (0,0) .. controls (3.31,-0.3) and (6.95,-1.4) .. (10.93,-3.29)(0,0) .. controls (3.31,0.3) and (6.95,1.4) .. (10.93,3.29)   ;

\draw    (72.5, 337) rectangle (407.5, 377)   ;
\draw    (239.75, 40) circle [x radius= 176.75, y radius= 20]  ;
\draw    (416.5,40) -- (415.5,427) ;

\draw    (63,40) -- (62,427) ;

\draw    (238.75, 427) circle [x radius= 176.75, y radius= 20]  ;
\draw   (425.5,219.37) .. controls (430.17,219.37) and (432.5,217.04) .. (432.5,212.37) -- (432.5,157.64) .. controls (432.5,150.97) and (434.83,147.64) .. (439.5,147.64) .. controls (434.83,147.64) and (432.5,144.31) .. (432.5,137.64)(432.5,140.64) -- (432.5,82.91) .. controls (432.5,78.24) and (430.17,75.91) .. (425.5,75.91) ;
\draw   (494.5,325.92) .. controls (499.17,325.92) and (501.5,323.59) .. (501.5,318.92) -- (501.5,210.16) .. controls (501.5,203.49) and (503.83,200.16) .. (508.5,200.16) .. controls (503.83,200.16) and (501.5,196.83) .. (501.5,190.16)(501.5,193.16) -- (501.5,81.41) .. controls (501.5,76.74) and (499.17,74.41) .. (494.5,74.41) ;
\draw   (564.5,387.92) .. controls (569.17,387.92) and (571.5,385.59) .. (571.5,380.92) -- (571.5,240.42) .. controls (571.5,233.75) and (573.83,230.42) .. (578.5,230.42) .. controls (573.83,230.42) and (571.5,227.09) .. (571.5,220.42)(571.5,223.42) -- (571.5,79.92) .. controls (571.5,75.25) and (569.17,72.92) .. (564.5,72.92) ;
\draw    (123.5, 70) rectangle (346.5, 110)   ;

\draw (229,142) node  [align=left] {Keys and Unit-Length Normalized Vectors};
\draw (236,208) node  [align=left] {SQLite Index over Keys};
\draw (241,248) node  [align=left] {Character N-Grams Enumerated for all Keys};
\draw (237,316) node  [align=left] {SQLite Full-Text Search Index over all N-Grams};
\draw (240,358) node  [align=left] {LZ4 Compressed Annoy {\fontfamily{pcr}\selectfont mmap} Index for all Vectors};
\draw (239,41) node  [align=left] {\textbf{SQLite Database}};
\draw (459,149.26) node [rotate=-90] [align=left] {Light};
\draw (528,203) node [rotate=-90] [align=left] {Medium};
\draw (598,233.97) node [rotate=-90] [align=left] {Heavy};
\draw (235,91) node  [align=left] {Format Settings and Metadata};

\end{tikzpicture}
\end{adjustbox}
}
\caption{\label{file-format} Structure of the ``.magnitude" file format and its Light, Medium, and Heavy variants. }
\end{figure}
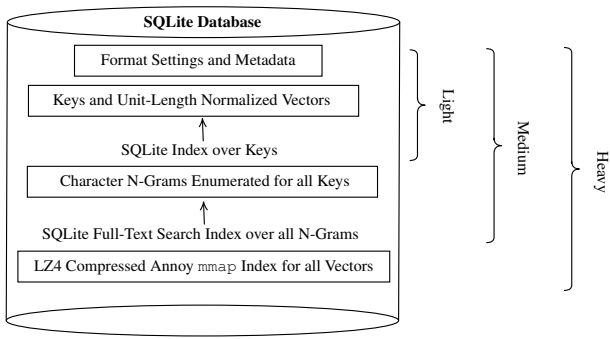

\paragraph{Converter}
\label{sec:converter}
The software includes a command-line converter utility for converting word2vec (``.bin", ``.txt"), GloVe (``.txt"), fastText (``.vec"), or ELMo (``.hdf5") files to Magnitude files. They can be converted with the command:

\begin{bash}
python -m pymagnitude.converter 
  -i "./vecs.(bin|txt|vec|hdf5)"
  -o "./vecs.magnitude"
\end{bash}

\noindent The input format will automatically be determined by the extension and the contents of the input file. When the vectors are converted, they will also be unit-length normalized. This conversion process only needs to be completed once per model. After converting, the Magnitude file format is static and it will not be modified or written to in order to make concurrent read access safe.

By default, the converter builds a Medium ``.magnitude" file. Passing the {\small\tt{-s}} flag will turn off encoding of subword information, and result in a Light flavored file. Passing the {\small\tt{-a}} flag will turn on building the Annoy approximate similarity index, and result in a Heavy flavored file. Refer to the documentation\footnote{\url{https://github.com/plasticityai/magnitude\#file-format-and-converter}} for more information about conversion configuration options.

\paragraph{Quantization}
The converter utility accepts a {\small\tt{-p <PRECISION>}} flag to specify the decimal precision to retain. Since underlying values are stored as integers instead of floats, this is essentially quantization\footnote{\url{https://www.tensorflow.org/performance/quantization}} for smaller model footprints. Lower decimal precision will create smaller files, because SQLite can store integers with either 1, 2, 3, 4, 6, or 8 bytes.\footnote{\url{https://www.sqlite.org/datatype3.html}} Regardless of the precision selected, the library will create {\small\tt{numpy.float32}} vectors. The datatype can be changed by passing {\small\tt{dtype=numpy.float16}} to the {\small\tt{Magnitude}} constructor.



\section{Conclusion}
Magnitude is a new open source Python library and file format for vector embeddings.  It makes it easy to integrate embeddings into applications and provides a single interface and configuration that is suitable for both development and production workloads. The library and file format also enable novel features like OOV handling that allow models to be more robust to noisy data. The simple interface, ease of use, and speed of the library, compared to other utilities like Gensim, will enable use by beginners to NLP and individuals in educational environments, such as university NLP and AI courses.

Pre-trained word embeddings have been widely adopted in NLP. Researchers in computer vision have started using pre-trained vector embedding models like Deep1B~\cite{babenko2016efficient} for images. The Magnitude library intends to stay agnostic to various domains, instead providing a generic key-vector store and interface that is useful for all domains and for research that crosses the boundaries between NLP and vision~\cite{Hewitt-et-al:2018:ACL}.


\section{Software and Data}
We release the Magnitude package under the permissive MIT open source license. The full source code and pre-converted ``.magnitude" models are on  GitHub. The full documentation for all classes, methods, and configurations of the library can be found at \url{https://github.com/plasticityai/magnitude}, along with example usage and tutorials.

We have pre-converted several popular embedding models (Google News word2vec, Stanford GloVe, Facebook fastText, AI2 ELMo) to ``.magnitude" in all its variants (Light, Medium, and Heavy).  You can download them from \url{https://github.com/plasticityai/magnitude#pre-converted-magnitude-formats-of-popular-embeddings-models}.

 \section*{Acknowledgments}
 We would like to thank Erik Bernhardsson for the useful feedback on integrating Annoy indexing into Magnitude and thank the numerous contributors who have opened issues, reported bugs, or suggested technical enhancements for Magnitude on GitHub.
 
This material is funded in part by DARPA under grant number HR0011-15-C-0115 (the LORELEI program) and by NSF SBIR Award \#IIP-1820240. The U.S. Government is authorized to reproduce and distribute reprints for Governmental purposes. The views and conclusions contained in this publication are those of the authors and should not be interpreted as representing official policies or endorsements of DARPA, the NSF, and the U.S. Government. This work has also been supported by the French National Research Agency under project ANR-16-CE33-0013.

\bibliography{emnlp2018}
\bibliographystyle{acl_natbib_nourl}

\appendix
\onecolumn
\section{Benchmark Comparisons}
\label{sec:benchmarks}
All benchmarks\footnote{\url{https://github.com/plasticityai/magnitude/blob/master/tests/benchmark.py}} were performed on the Google News pre-trained word vectors, ``GoogleNews-vectors-negative300.bin"~\cite{DBLP:journals/corr/abs-1301-3781} for Gensim and on the ``GoogleNews-vectors-negative300.magnitude"\footnote{\url{http://magnitude.plasticity.ai/word2vec+approx/GoogleNews-vectors-negative300.magnitude}} for Magnitude, with a MacBook Pro (Retina, 15-inch, Mid 2014) 2.2GHz quad-core Intel Core i7 @ 16GB RAM on a SSD over an average of trials where feasible. We are explicitly not using Gensim's memory-mapped native format as it requires extra configuration from the developer and is not provided out of the box from Gensim's data repository~\footnote{\url{https://github.com/RaRe-Technologies/gensim-data}}.

\newcolumntype{M}[1]{>{\centering\arraybackslash}m{#1}}
\newcolumntype{L}[1]{>{\raggedright\arraybackslash}p{#1}}
\begin{table*}[!htbp]
\begin{center}
\begin{minipage}{\textwidth}
\small
{\renewcommand{\arraystretch}{1.15}
\begin{tabular}{@{}L{7cm}@{}|@{}M{2.15cm}@{}@{}M{2.15cm}@{}@{}M{2.15cm}@{}@{}M{2.15cm}@{}}
\hline \bf Metric & \bf ~~~~Gensim\newline\cite{rehurek_lrec}   & \bf ~~~Magnitude\newline Light & \bf ~~~Magnitude\newline Medium & \bf ~~~~Magnitude\newline Heavy \\ \hline
Initial load time                                                                                                                                      & 70.26s & \textbf{0.7210s}          & ---~\footnote{\label{previousvalue} Denotes the same value as the previous column.}   & ---~\footref{previousvalue} \\\hline
Cold single key query                                                                                                                                  & \textbf{0.0001s} & \textbf{0.0001s}          & ---~\footref{previousvalue}  & ---~\footref{previousvalue} \\\hline
Warm single key query \\\small{\it{(same key as cold query)}}                                                                                     & 0.0044s & \textbf{0.00004s}         & ---~\footref{previousvalue}  & ---~\footref{previousvalue} \\\hline
Cold multiple key query \\\small{\it{(n=25)}}                                                                                                     & 3.0050s & \textbf{0.0442s}          & ---~\footref{previousvalue}  & ---~\footref{previousvalue} \\\hline
Warm multiple key query \\\small{\it{(n=25) (same keys as cold query)}}                                                                           & \textbf{0.0001s} & \textbf{0.00004s}         & ---~\footref{previousvalue}  & ---~\footref{previousvalue} \\\hline
First {\small\tt{most\_similar}} search query \\\small{\it{(n=10) (worst case)}}                                                                              & \textbf{18.493s} & 247.05s               & ---~\footref{previousvalue}  & ---~\footref{previousvalue} \\\hline
First {\small\tt{most\_similar}} search query \\\small{\it{(n=10) (average case) (w/ disk persistent cache)}}                                                 & 18.917s & \textbf{1.8217s}          & ---~\footref{previousvalue}  & ---~\footref{previousvalue} \\\hline
Subsequent {\small\tt{most\_similar}} search \\\small{\it{(n=10) (different key than first query)}}                                                           & 0.2546s & \textbf{0.2434s}          & ---~\footref{previousvalue}  & ---~\footref{previousvalue} \\\hline
Warm subsequent {\small\tt{most\_similar}} search \\\small{\it{(n=10) (same key as first query)}}                                                             & 0.2374s & \textbf{0.00004s}         & \textbf{0.00004s}        & \textbf{0.00004s}       \\\hline
First {\small\tt{most\_similar\_approx}} search query \\\small{\it{(n=10, effort=1.0) (worst case)}}                                                           & N/A~\footnote{\label{approx} Gensim does support approximate similarity search, but not out of the box as the index must be built manually with {\small\tt{gensim.similarities.index}} first which is a slow operation.} & N/A                   & N/A                  & \textbf{29.610s}        \\\hline
First {\small\tt{most\_similar\_approx}} search query \\\small{\it{(n=10, effort=1.0) (average case) (w/ disk persistent cache)}}                              & N/A & N/A                   & N/A                  & \textbf{0.9155s}        \\\hline
Subsequent {\small\tt{most\_similar\_approx}} search \\\small{\it{(n=10, effort=1.0) (different key than first query)}}                                        & N/A & N/A                   & N/A                  & \textbf{0.1873s}        \\\hline
Subsequent {\small\tt{most\_similar\_approx}} search \\\small{\it{(n=10, effort=0.1) (different key than first query)}}                                        & N/A & N/A                   & N/A                  & \textbf{0.0199s}        \\\hline
Warm subsequent {\small\tt{most\_similar\_approx}} search \\\small{\it{(n=10, effort=1.0) (same key as first query)}}                                          & N/A & N/A                   & N/A                  & \textbf{0.00004s}       \\\hline
File size                                                                                                                                              &  \textbf{3.64GB} & 4.21GB                & 5.29GB               & 10.74GB             \\\hline
Process memory (RAM) utilization                                                                                                                       & 4.875GB & \textbf{18KB}             & ---~\footref{previousvalue}  & ---~\footref{previousvalue} \\\hline
Process memory (RAM) utilization after 100 key queries                                                                                                 & 4.875GB & \textbf{168KB}            & ---~\footref{previousvalue}  & ---~\footref{previousvalue} \\\hline
Process memory (RAM) utilization after 100 key queries + similarity search                                                                             & 8.228GB~\footnote{\label{unitdup} Gensim has an option to not duplicate unit-normalized vectors in memory, but still requires up to 8GB of memory allocation while processing, before dropping down to half the memory. Moreover, this option is not on by default.} & \textbf{342KB}~\footnote{\label{mmap} Magnitude uses {\small\tt{mmap}} to read from the disk, so the OS will still allocate pages of memory, when memory is available, in its file cache, but it can be shared between processes and is not managed within each process for extremely large files which is a performance win.} & ---~\footref{previousvalue}  & ---~\footref{previousvalue} \\
\end{tabular}
}
\end{minipage}
\end{center}
\caption{\label{metric-table} Benchmark comparisons between Gensim, Magnitude Light, Magnitude Medium, and Magnitude Heavy. }
\end{table*}

\end{document}